\documentclass{bioma} 

\usepackage[english]{babel}
\usepackage{amsmath}
\usepackage{amssymb}    

\usepackage{algorithm,algorithmic}
\usepackage{epsfig}
\usepackage{graphics,graphicx}

\usepackage{url}
\usepackage{listings}

\usepackage{cite}
\usepackage[table]{xcolor}
\usepackage{hyperref}

\normallatexbib

\newcommand{\longtitle}{A Particle Swarm Optimization hyper-heuristic for the Dynamic Vehicle Routing Problem}
\newcommand{\okulewicz}{Micha{\l} Okulewicz}
\newcommand{\mandziuk}{Jacek Ma{\'n}dziuk}
\newcommand{\okulewiczmail}{M.Okulewicz}
\newcommand{\mandziukmail}{J.Mandziuk}
\newcommand{\miniwutmail}{@mini.pw.edu.pl}
\newcommand{\miniwut}{Faculty of Mathematics and Information Science\\
Warsaw University of Technology, Warsaw, Poland}
\newcommand{\paperkeywords}{Dynamic Vehicle Routing Problem,Particle Swarm Optimization,Hyper-heuristic}

\begin{document}

\articletitle{\longtitle}

\author{\okulewicz}
\affil{\miniwut}
\email{\okulewiczmail\miniwutmail}

\author{\mandziuk}
\affil{\miniwut}
\email{\mandziukmail\miniwutmail}

\begin{abstract}
This paper presents a method for choosing a Particle Swarm Optimization based optimizer
for the~Dynamic Vehicle Routing Problem on the~basis
of the initially available data of a given problem instance.
The~optimization algorithm is chosen on the basis of a prediction made by a linear model
trained on that data and the relative results obtained by the optimization algorithms.
The achieved results suggest that such a~model can be used
in a hyper-heuristic approach as it improved the average results, obtained on the set
of benchmark instances, by choosing the appropriate algorithm in 82\% of significant cases.
Two leading multi-swarm Particle Swarm Optimization based algorithms for solving the Dynamic Vehicle Routing Problem are used as the basic optimization algorithms: Khouadjia's et al. Multi--Environmental Multi--Swarm Optimizer
and~authors' 2--Phase Multiswarm Particle Swarm Optimization.
\end{abstract}

\begin{keywords}
\paperkeywords
\end{keywords}

\section{Introduction}
Dynamic transportation problems have been considered in the literature
by Psaraftis \cite{Psaraftis1980130,DVRP:Reintroduction} since 1980. After the
introduction of a set benchmarks by Kilby \cite{DVRP:Study}
and Montemanni \cite{DVRP:Ants}, several meta-heuristic based
algorithms have been developed in order to solve the problem,
including Ant Colony Optimization (ACS) \cite{DVRP:Ants,DVRP:ACOLNS},
Genetic Algorithm (GA) \cite{DVRP:GA:TS,DVRP:GA2014},
Tabu Search (TS) \cite{DVRP:GA:TS}, and Particle Swarm Optimization (PSO) \cite{DVRP:MEMSO,DVRP:2MPSO}.

Although some of the works \cite{DVRP:GA:TS,DVRP:GA2014,DVRP:2PSO,DVRP:DAPSO} mention the features
of a~spatial distribution of the requests of a given DVRP instance (describing it as spatially uniform
or clustered), none of those methods directly use
the information about the known requests location
and volume. 
An initial non-parametric approach for using the information about
the available requests volumes in order
to generate artificial requests (to account for the unknown
ones) has been presented by the authors \cite{DVRP:MCTree}.

This paper proposes a hyper-heuristic method based on Multi--Envi-ronmental Multi--Swarm Optimizer (MEMSO) \cite{DVRP:DAPSO,DVRP:MAPSO,DVRP:MEMSO}
and 2--Phase Multiswarm Particle Swarm Optimization (2MPSO) \cite{DVRP:2PSO,DVRP:2MPSO,DVRP:MCTree} algorithms. The hyper-heuristic
uses the statistical data about the initially known set of requests in the given Dynamic Vehicle Routing Problem (DVRP).

The rest of the paper is organized as follows. Section \ref{sec:Okulewicz-dvrp} defines the~DVRP solved in this paper. Section \ref{sec:Okulewicz-memso.2mpso} introduces PSO and algorithms MEMSO and 2MPSO (both based on the PSO) used for optimizing DVRP.
Section \ref{sec:Okulewicz-hyperheuristic} presents
the hyper-heuristic approach for solving the~DVRP. Section \ref{sec:Okulewicz-results} gives experimental
setup and results obtained by the method. Finally, Section \ref{sec:Okulewicz-conclusions} concludes
the paper.

\section{Dynamic Vehicle Routing Problem}
\label{sec:Okulewicz-dvrp}
DVRP is a dynamic version of the generalization
of a Traveling Salesman Problem (TSP), called Vehicle Routing Problem (VRP).
In the VRP the goal is to optimize a total route for a fleet of vehicles with a limited capacity.
VRP has been introduced as a \emph{Truck dispatching problem} in 1959
by Dantzig and Ramser \cite{VRP:Definition}. After the technological advancement
of the vehicle tracking devices and development of the Geographical Information Systems
a notion of a DVRP has been reintroduced
by Psaraftis \cite{DVRP:Reintroduction}.
The problem has attracted more attention after a set of static VRP benchmarks
of Christofides \cite{VRP:ChristofidesBeasley84}, Fisher \cite{VRP:FisherJakumar81} and Taillard \cite{VRP:Taillard93} 
has been customized for the DVRP by Kilby et al. in 1998~\cite{DVRP:Study} and Montemanni et al. in 2005~\cite{DVRP:Ants}.

In this paper a most common variant of the DVRP is solved \cite{DVRP:Review2011},
sometimes referred to as a VRP with Dynamic Requests (VRPwDR) \cite{DVRP:DAPSO}.
In this variant a homogeneous fleet of vehicles (identical capacity $c \in \mathbb{R}$ and speed $sp \in \mathbb{R}$) is considered.
There is also an additional constraint, that the vehicle may operate only
during a \emph{working day} defined by the opening hours of its depot.
During that working day a fleet of $m$ vehicles must serve (visit) a set of $n$ requests.
Each request is defined by a~location $l_i \in \mathbb{R}^2$,
an amount of cargo $s_i$ ($0 \le s_i \leq c$) to be delivered 
and an amount of time $u_i \in \mathbb{R}$ it takes to provide a service at that location (unload that cargo).
The dynamic nature of that optimization problem comes from the fact that new request may
arrive during the working day. The period of time during which new requests are accepted
is limited by a \emph{cut--off time} parameter $T_{co}$. In all the benchmarks considered
in this paper there is one depot for all the vehicles and $T_{co}$ is equal to a half
of the working day.

In the mathematical sense, the DVRP is a problem of finding a set of location
permutations resulting in a shortest total path length under the time
and capacity constrains for each of the permutations in which all the locations
are visited exactly once. The exact mathematical formulation of the problem can be
found in \cite{DVRP:GA2014,DVRP:2MPSO,DVRP:MCTree}.

\section{MEMSO and 2MPSO Algorithms}
\label{sec:Okulewicz-memso.2mpso}
In this section two most successful approaches to solving DVRP:
Khouadjia's~et~al. MEMSO
and~authors' 2MPSO
are presented. Those two methods are the base algorithms
for the hyper--heuristic approach proposed in this paper.

Both, the MEMSO and the 2MPSO, use PSO
as their base meta-heuristic and 2-OPT \cite{TSP:2OPT} as a route optimization heuristic.
In both methods the working day is
divided into discrete number of time slices, with the instance
of the DVRP problem "frozen" within each time slice.
Therefore, each method solves a series of dependent static VRP
instances during the optimization process.
The difference between PSO applications, encoding of the problem
and knowledge transfer in the MEMSO and the 2MPSO methods,
together with a brief description of the PSO and 2-OPT algorithms,
are presented and discussed in this section.

\subsection{Particle Swarm Optimization}
PSO algorithm is an iterative population based
continuous optimization meta-heuristic approach utilizing
the concept of~Swarm Intelligence.
The algorithm has been introduced by Kennedy and Eberhart in 1995 \cite{PSO:Introduction}
and has been  further developed and studied by other researchers \cite{PSO:Modified,PSO:Params,PSO:Convergence}.

During the optimization process PSO maintains a set of fitness function solutions
(called particles). Each particle has its own location $x \in \mathbb{R}^n$ (an $n-$dimensional fitness function solution proposal),
velocity $v$ (a solution change vector), a set of neighbors $N$ (particles which solutions it can observe),
a memory of the best observed solution $x_N^{(BEST)}$ and a memory of the best visited solution $x_i^{(BEST)}$.

In each iteration $t$ the location vector $x$ and velocity vector $v$ of $i$th particle are changed in the following way:
\begin{equation}
	x_{i,t} = x_{i,t-1} + v_{i,t-1}
	\label{eq:Okulewicz-position}
\end{equation}

\begin{equation}
	v_{i,t} = \omega v_{i,t-1} + u_1c_1(x_i^{(BEST)} - x_{i,t}) + u_2c_2(x_N^{(BEST)} - x_{i,t})
\end{equation}
Where $\omega$ denotes an inertia factor, $c_1$ and $c_2$ are personal and global attraction factors,
$u_1$  and $u_2$ follow the uniform $n-$dimensional distribution on $[0,1]^n$.

\subsection{2--OPT Algorithm}
\begin{figure}[h!]
\includegraphics[width=\textwidth]{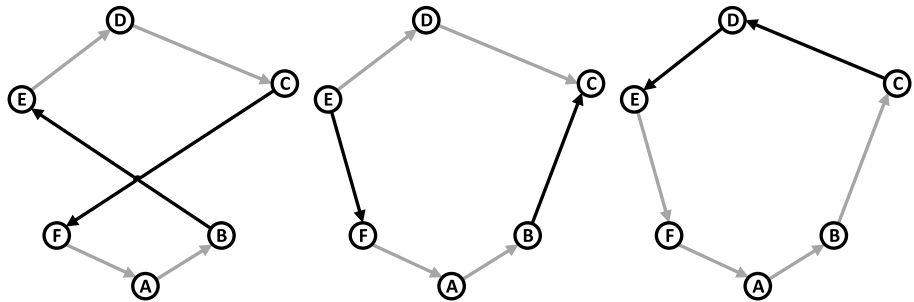}
\caption{Depiction of a 2--OPT algorithm optimization process over a sample directed cycle.
While considering edges BE and CF the algorithm observed that changing them to BC and EF
results in a shorter route. After swapping the ends of the considered edges
a final step of reversing the direction of the E-D-C path is performed.
\label{fig:Okulewicz-2OPT}}
\end{figure}

2--OPT has been introduced as a heuristic algorithm for solving the TSP in 1958 \cite{TSP:2OPT}.
Its most distinctive feature is the ability to remove the entanglement of routes.
The algorithm operates by iterating over all the pairs of edges of a given route
and checking the possibility of optimizing the length of route by swapping the ends of those edges.
An example of a single step of the algorithm on a sample directed cycle is presented in~Fig.~\ref{fig:Okulewicz-2OPT}.

2--OPT may be used directly in the VRP variants for optimizing the length of route
of a single vehicle.

\subsection{Multi--Environmental Multi--Swarm Optimizer}
MEMSO~\cite{DVRP:MEMSO} algorithm uses PSO to optimize division of the requests among the vehicles.
The vehicles' routes within those divided sets are created by a greedy insertion
and optimized by the 2-OPT algorithm.
The fitness function value is the total length of those routes.

MEMSO uses a discrete encoding of the requests division.
The solution is an integer vector representing the requests
and the values in the vector are the vehicles' identifiers.
Therefore, the PSO algorithm is changed in such a way
that a velocity is a vector in $\lbrace 1,2,\ldots,m \rbrace^n$
and addition in eq.~(\ref{eq:Okulewicz-position}) is performed
in ${\mathbb{Z}_m}^n$ space (where $m$ is a number of vehicles
and $n$ a number of requests).

The knowledge about the current solution is transferred between
subsequent time slices by adapting the whole population.
For each of the particles the already served and decisively assigned requests
are blocked from being changed and new requests are inserted in~a~greedy way
into the solution vectors.

\subsection{2-Phase Multi-swarm Particle Swarm Optimization}
2MPSO~\cite{DVRP:2MPSO} algorithm uses separate PSO instances
to optimize both the division of the requests
and their order (in two subsequent optimization phases, hence the name of the method).
In the division optimization phase an approach similar
to the MEMSO's is used in order to evaluate the total length of routes
achieved from the optimized division.
The routes are optimized with 2-OPT algorithm from the initially random ordering.
In the route optimization phase each of the vehicles is optimized separately
and the length of route of the given vehicle is used as a fitness function value.

Instead of customizing the PSO for a~discrete problem, 2MPSO follows a continuous optimization approach to applying the PSO algorithm, in contrast with MEMSO.
The~division of requests among the~vehicles is solved as a clustering task,
with a number of clusters per vehicle $k$ being the parameter of the method.
Therefore, the PSO particle for the first phase optimizer
is a sequence of requests clusters centers flattened into
a vector in $\mathbb{R}^{2k\hat{m}}$ space (where $\hat{m}$
is the estimated number of vehicles necessary to serve the requests).
Particle in the PSO instance optimizing the route of the vehicle
is a sequence of requests ranks. Therefore, it is a vector in
$\mathbb{R}^{n_i}$ (where $n_i$ is the number of requests assigned
to $i$th vehicle).

The knowledge is transferred between subsequent time slices
in a form of cluster centers vector generating
a division of requests and a set of requests rank vectors
for the routes imposed by requests division.
The transferred solution is expanded by a new random
cluster center if more vehicles seem to be necessary 
and the initial ranks of the new requests are also
initialized at random.

\section{Hyper-heuristic Approach}
\label{sec:Okulewicz-hyperheuristic}

\begin{figure}[!ht]
	\centering
	\includegraphics[width=\textwidth]{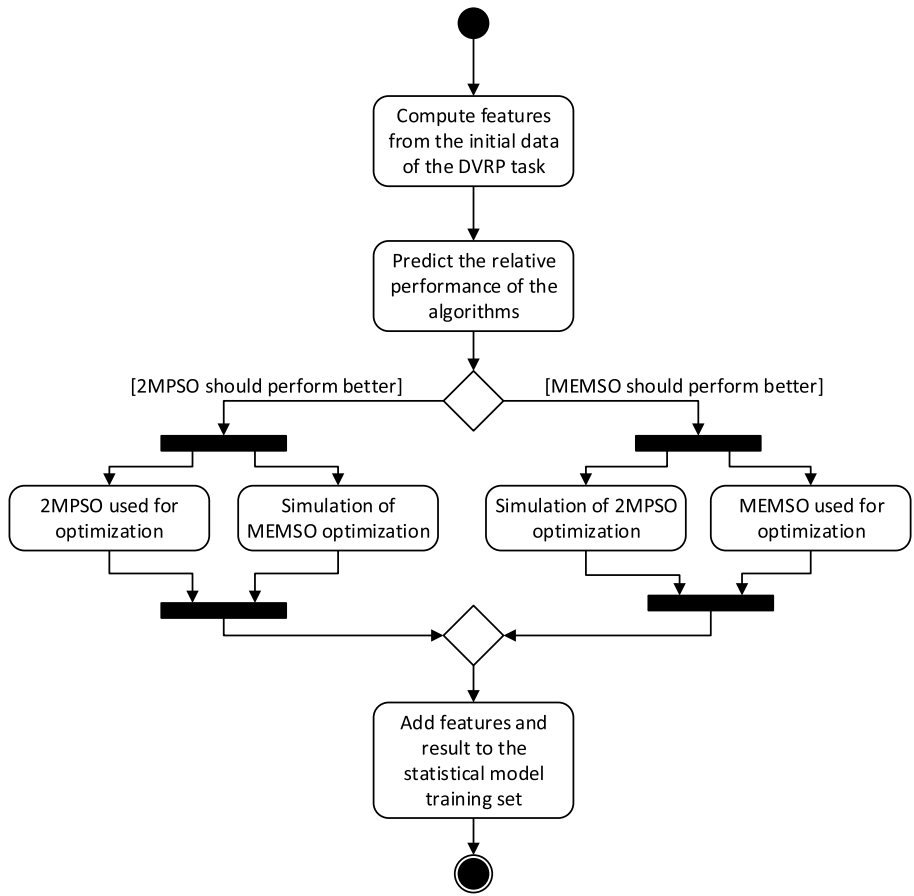}
	\caption{The activity diagram presenting a single run of a 2MPSO and MEMSO based
	hyper-heuristic.
	\label{fig:Okulewicz-hyperactivity}}
\end{figure}

Although 2MPSO outperforms MEMSO by 2.22\% on average on the Kilby's \cite{DVRP:Study}
and Montemanni's \cite{DVRP:Ants} sets of benchmarks (see Table \ref{tab:Okulewicz-results}),
it might be limited by its clustering approach (or needs an excessively large $k$)
for some particular DVRP tasks.

For that reason, the authors propose a hyper-heuristic approach \cite{burke2003hyper}
in which a statistical model
might be incrementally trained for choosing the algorithm, which seems to be most suitable
for a given DVRP instance.
A single run of such hyper-heuristic (solving a single DVRP task)
is depicted in Fig.~\ref{fig:Okulewicz-hyperactivity}.
Please observe in the activity diagram, that only the chosen algorithm is used
to provide an actual output to some external decision support system. The other algorithm is run
only to gather the data and its result is used to tune the prediction model.
Please also note, that the choice of the optimization algorithm is done only
at the beginning of the optimization process.
The prediction model is trained on the results gathered through a subsequent runs on different DVRP tasks.

The economic cost of running such a hyper-heuristic, in comparison with a single algorithm optimization,
would be slightly more than doubled. The doubling comes from the fact, that it is necessary
to make a~similar amount of computations by two algorithms in order to get comparable
results for the performance prediction model. The additional overhead is a result
of computations needed for getting statistics from the set of requests
and creating a prediction model over the already computed cases.
Although the economic cost of proposed approach would definitely be larger
than that of a single algorithm,
the overhead for the computations necessary for getting solutions
during daytime operations would be negligible.
The additional operations during the daytime
would consist of computing the statistics from the initial state
of requests set and providing them to the prediction model.
The run of the second algorithm and the training of the prediction model 
might be done during the nighttime.

This section presents the statistics computed from the benchmark problems
and authors' approach to creating a linear model, based on those statistics,
predicting the relative MEMSO and 2MPSO algorithms performance.

\subsection{Benchmark Characteristics}
\label{sec:Okulewicz-characteristics}
\begin{table}[!ht]
	\caption{Values of the input features computed for the initially known sets of requests from Kilby's and Montemanni's benchmarks}
	\label{tab:Okulewicz-features}
	\resizebox{\textwidth}{!}{
	\begin{tabular}{l|rrrrrrrrrr}
	Name & $\mu_x$ & $sd_x$ & $skew_x$& $\mu_y$ & $sd_y$ & $skew_y$& $\mu_s$ & $sd_s$ & $skew_s$ & $nc$ \\\hline
	c50 & 0.53 & 0.32 & -0.18 & 0.51 & 0.33 &  0.08 & 0.09 & 0.04 & 0.58 & 2.00 \\
c75 & 0.51 & 0.30 & -0.08 & 0.41 & 0.29 &  0.27 & 0.12 & 0.06 & 0.34 & 4.00 \\
c100 & 0.48 & 0.27 &  0.35 & 0.46 & 0.27 &  0.53 & 0.08 & 0.05 & 0.71 & 1.00 \\
c100b & 0.68 & 0.17 &  0.48 & 0.56 & 0.26 &  0.31 & 0.09 & 0.05 & 1.39 & 0.00 \\
c120 & 0.38 & 0.32 &  0.84 & 0.67 & 0.24 &  0.20 & 0.06 & 0.03 & 1.46 & 0.67 \\
c150 & 0.49 & 0.27 &  0.06 & 0.45 & 0.24 &  0.28 & 0.08 & 0.04 & 0.78 & 6.00 \\
c199 & 0.52 & 0.26 & -0.04 & 0.47 & 0.24 &  0.19 & 0.09 & 0.04 & 0.52 & 9.00 \\\hline
f71 & 0.44 & 0.23 & -0.02 & 0.72 & 0.18 & -0.17 & 0.04 & 0.05 & 1.87 & 0.00 \\
f134 & 0.61 & 0.29 & -0.47 & 0.42 & 0.22 &  0.70 & 0.07 & 0.11 & 2.35 & 1.50 \\\hline
tai75a & 0.27 & 0.18 &  1.39 & 0.47 & 0.20 &  1.22 & 0.11 & 0.16 & 2.27 & 3.00 \\
tai75b & 0.69 & 0.25 & -1.14 & 0.42 & 0.22 & -0.69 & 0.11 & 0.16 & 1.24 & 1.00 \\
tai75c & 0.43 & 0.16 & -0.93 & 0.49 & 0.25 & -0.02 & 0.11 & 0.15 & 2.30 & 3.00 \\
tai75d & 0.43 & 0.30 &  0.55 & 0.49 & 0.29 & -0.23 & 0.09 & 0.13 & 1.55 & 0.25 \\
tai100a & 0.39 & 0.30 &  0.61 & 0.57 & 0.25 & -0.27 & 0.11 & 0.18 & 2.04 & 1.00 \\
tai100b & 0.46 & 0.29 &  0.46 & 0.46 & 0.22 &  0.56 & 0.12 & 0.16 & 1.64 & 1.00 \\
tai100c & 0.63 & 0.27 & -0.91 & 0.50 & 0.20 & -1.10 & 0.10 & 0.14 & 1.34 & 5.00 \\
tai100d & 0.44 & 0.23 & -0.39 & 0.46 & 0.21 &  0.06 & 0.10 & 0.17 & 2.37 & 2.00 \\
tai150a & 0.47 & 0.29 &  0.17 & 0.55 & 0.31 & -0.36 & 0.09 & 0.14 & 2.32 & 1.67 \\
tai150b & 0.61 & 0.23 &  0.13 & 0.52 & 0.24 &  0.81 & 0.09 & 0.14 & 2.30 & 6.00 \\
tai150c & 0.47 & 0.23 &  0.46 & 0.61 & 0.14 & -0.06 & 0.09 & 0.13 & 1.80 & 6.00 \\
tai150d & 0.47 & 0.34 & -0.05 & 0.67 & 0.23 & -0.88 & 0.12 & 0.16 & 1.41 & 0.14 \\

	\end{tabular}
	}
\end{table}

In order to create a set of features, allowing to discriminate between
different benchmark problems, the authors have computed a
set of statistics for each of the benchmarks. Those statistics may be divided into 2 groups:
\begin{itemize}
	\item requests set spatial features,
	\item requests set volume features.
\end{itemize}

\noindent Within those groups, the following statistics have been computed:
\begin{itemize}
\item {Spatial features:
\begin{itemize}
	\item $\mu_x$, $\mu_y$ - mean locations along coordinate system axes,
	\item $sd_x$, $sd_y$ - standard deviation of locations along coordinate system axes, 
	\item $skew_x$, $skew_y$ - standardized skewness of locations along coordinate system axes, 
	\item $k_{gap}$ - gap statistic (estimated optimal number of clusters) for requests locations,
\end{itemize}}
\item {Volume features:
\begin{itemize}
	\item $\mu_s$ - mean volume,
	\item $sd_s$ - standard deviation of volume,
	\item $skew_s$ - skewness of volume,
	\item $m_v$ - minimum number of vehicles necessary to load all the requests.
\end{itemize}}
\end{itemize}

In order to make the features comparable between different benchmarks,
the spatial locations have been mapped to the $[0,1] \times [0,1]$ plain,
requests volume has been divided by vehicles capacity and $k_{gap}$
has been combined with the $m_v$ in the following way:
\begin{equation*}
	nc = |1-\dfrac{m_v}{k_{gap}}|
\end{equation*}
The larger values of $nc$ suggest that the requests might not be easily
divided among the vehicles.

The results of computing those features on the Kilby's and Montemanni's benchmark set
for the \emph{a priori} available requests are presented in~Table~\ref{tab:Okulewicz-features}.

\subsection{Prediction Model}

\begin{lstlisting}[language=sh,float=!ht,frame=tb,caption={Model trained on the full data set, with the variable selection using Akaike Information Criterion, presented as an R output.},label=fig:Okulewicz-model]
Coefficients:
            Estimate Std. Error t value Pr(>|t|)    
(Intercept)  1.42899    0.15676   9.116 9.64e-07 ***
sd_x         1.23119    0.26551   4.637 0.000573 ***
mu_y        -1.11031    0.16881  -6.577 2.62e-05 ***
sd_y        -1.56756    0.28836  -5.436 0.000151 ***
skew_y      -0.04576    0.02481  -1.844 0.089977 .  
mu_s         2.72795    1.25435   2.175 0.050362 .  
sd_s        -3.39358    0.71918  -4.719 0.000498 ***
skew_s       0.21507    0.04592   4.683 0.000529 ***
---
Signif. codes:  0 '***' 0.001 '**' 0.01 '*' 0.05 '.'
Residual standard error:
                  0.03915 on 12 degrees of freedom
Multiple R-squared: 0.86, Adjusted R-squared: 0.7764 
F-statistic: 10.43 on 7 and 12 DF, p-value: 0.000286
\end{lstlisting}

In order to chose a proper optimization algorithm, the authors propose a simple linear regression model.

To train such a model, for predicting the proper optimization algorithm
for a given benchmark,
the statistics presented in Tab.~\ref{tab:Okulewicz-features} has been selected as an input variables
and a ratio between the average MEMSO and the average 2MPSO result has been chosen
as an output variable. Such approach is justified by the fact that it is more important to choose
a proper algorithm when the difference between different algorithms performance is significant.

Fitted linear model is used in a following way.
The subset of the statistics to be computed, described in Section~\ref{sec:Okulewicz-characteristics}, is identified with the usage of a variable selection method (cf. Listing \ref{fig:Okulewicz-model}).
When a new DVRP instance is considered, that subset of statistics, forming the input variables for the model,
is computed over the requests of that instance. If the model predicts the relative result
of MEMSO to 2MPSO to be less than $1$, MEMSO is chosen as the optimization algorithm,
otherwise 2MPSO is chosen (see Fig.~\ref{fig:Okulewicz-hyperactivity}).

\section{Results}
\label{sec:Okulewicz-results}

\begin{table}[!ht]
	\caption{Results of predicting the optimization algorithm from a leave-one-out cross-validation experiment.
	The table presents the minimum and the average values achieved by MEMSO and 2MPSO algorithms,
	marking the significantly better average results with a gray background.
	The Hyper-heuristic column presents which algorithm has been chosen by a linear model,
	whether the choice has been appropriate and how much has been gained (or lost) with that choice
	of algorithm.
	For the benchmark instances with significant difference between average results of MEMSO and 2MPSO
	the gain has been marked with a gray background and a loss with a light-gray background.
	\label{tab:Okulewicz-results}}
	\centering
	\resizebox{\textwidth}{!}{
	\begin{tabular}{l|rr|rr|ccr}
	& \multicolumn{2}{c|}{2MPSO} & \multicolumn{2}{c|}{MEMSO} & \multicolumn{3}{c}{Hyper-heuristic} \\
	Name & min  & avg  & min & avg & Chosen & T/F & Gain \\\hline
	c50 &   583.09 &   618.59 & \textbf{577.60} & \cellcolor{gray!25}\textbf{592.95} & MEMSO & T & \cellcolor{gray!25}4.15\% \\
c75 &   \textbf{904.83} &   \cellcolor{gray!25}\textbf{946.85} &   928.53 &   962.54 & 2MPSO & T & \cellcolor{gray!25}1.63\% \\
c100 &   \textbf{926.10} &   \textbf{966.27} &   949.83 &   968.92 & MEMSO & F & -0.27\% \\
c100b &   \textbf{830.58} &   \textbf{875.47} &   864.19 &   878.81 & MEMSO & F & -0.38\% \\
c120 &  \textbf{1061.84} &  \cellcolor{gray!25}\textbf{1176.38} &  1164.63 &  1284.62 & 2MPSO & T & \cellcolor{gray!25}8.43\% \\
c150 &  \textbf{1132.12} &  \cellcolor{gray!25}\textbf{1208.60} &  1274.33 &  1327.24 & 2MPSO & T & \cellcolor{gray!25}8.94\% \\
c199 &  \textbf{1371.61} &  \cellcolor{gray!25}\textbf{1458.01} &  1600.57 &  1649.17 & 2MPSO & T & \cellcolor{gray!25}11.59\% \\\hline
f71 &   302.50 &   319.01 &   \textbf{283.43} &   \cellcolor{gray!25}\textbf{294.85} & 2MPSO & F & \cellcolor{gray!12}-7.57\% \\
f134 & \textbf{11944.86} & \cellcolor{gray!25}\textbf{12416.65} & 14814.10 & 16083.82 & 2MPSO & T & \cellcolor{gray!25}22.80\% \\\hline
tai75a &  \textbf{1721.81} &  1846.03 &  1785.11 &  \textbf{1837.00} & 2MPSO & F & -0.49\% \\
tai75b &  1418.82 &  1451.92 &  \textbf{1398.68} &  \cellcolor{gray!25}\textbf{1425.80} & MEMSO & T & \cellcolor{gray!25}1.80\% \\
tai75c &  \textbf{1456.90} &  1560.68 &  1490.32 &  \cellcolor{gray!25}\textbf{1532.45} & MEMSO & T & \cellcolor{gray!25}1.81\% \\
tai75d &  1445.58 &  1481.25 &  \textbf{1342.26} &  \cellcolor{gray!25}\textbf{1448.19} & MEMSO & T & \cellcolor{gray!25}2.23\% \\
tai100a &  2211.30 &  2327.20 &  \textbf{2170.54} &  \cellcolor{gray!25}\textbf{2213.75} & MEMSO & T & \cellcolor{gray!25}4.87\% \\
tai100b &  \textbf{2052.54} &  \cellcolor{gray!25}\textbf{2131.91} &  2093.54 &  2190.01 & 2MPSO & T & \cellcolor{gray!25}2.65\% \\
tai100c &  \textbf{1465.06} &  \cellcolor{gray!25}\textbf{1519.44} &  1491.13 &  1553.55 & 2MPSO & T & \cellcolor{gray!25}2.20\% \\
tai100d &  \textbf{1722.16} &  \cellcolor{gray!25}\textbf{1808.67} &  1732.38 &  1895.42 & 2MPSO & T & \cellcolor{gray!25}4.58\% \\
tai150a &  3367.55 &  3537.81 &  \textbf{3253.77} &  \cellcolor{gray!25}\textbf{3369.48} & 2MPSO & F & \cellcolor{gray!12}-4.76\% \\
tai150b &  2911.22 &  3033.83 &  \textbf{2865.17} &  \cellcolor{gray!25}\textbf{2959.15} & 2MPSO & F & \cellcolor{gray!12}-2.46\% \\
tai150c &  2510.51 &  \cellcolor{gray!25}\textbf{2579.72} &  \textbf{2510.13} &  2644.69 & 2MPSO & T & \cellcolor{gray!25}2.46\% \\
tai150d &  2893.54 &  \textbf{2992.53} &  \textbf{2872.80} &  3006.88 & MEMSO & F & -0.48\% \\

	\end{tabular}
	}
\end{table}

To test the proposed approach a leave-one-out cross-validation experiment with a linear model
predicting the relative performance of the MEMSO to 2MPSO  has been performed.
In each fold a linear model has been built on information from initial sets of~requests from 20 out of 21
benchmark problems. The relative algorithm performance has been computed as a ratio
of the average results obtained by the MEMSO and 2MPSO algorithms.
Such approach simulated a performance of~hyper-heuristic
optimizing and training the model on some number of~subsequently computed DVRP tasks.
The average results were computed over $30$ runs of MEMSO per benchmark and $20$ runs of 2MPSO per benchmark. The order of the total fitness function evaluations budget for each of the algorithms run has been equal to $10^6$.
The details about parameter setting for MEMSO and 2MPSO experiments can be found in \cite{DVRP:MEMSO} and \cite{DVRP:2MPSO}, respectively.

Additionally, in the model training phase some of the input variables have been removed from the model
in a step-wise mode, with the usage of Akaike Information Criterion, in order to create a more
general predictor. As an example, results of applying such procedure, to the model trained
on all 21 of the benchmarks, are presented as a listing of R output in Listing~\ref{fig:Okulewicz-model}.
It can be observed (from the corresponding $p$-values) that both the spatial
and the volume related features have been selected as informative ones.

The results from the cross-validation experiment are given in Table~\ref{tab:Okulewicz-results}.
The correct algorithm (the one with a better average result) has been chosen in 14 out of 21 cases (all of them with
statistically significant difference between average algorithms results, verified by a $t$-test).
Wrong predictions, that have been made for the other 7 cases, have resulted in a significant loss of results quality
only for 3 of them (\texttt{f71}, \texttt{tai150a}, and \texttt{tai150b}).
Therefore, the linear model achieved $82\%$ accuracy for the subset of $17$ benchmarks with significant differences
in the algorithms average results (with $67\%$ accuracy over the whole set of benchmarks).

\section{Conclusions}
\label{sec:Okulewicz-conclusions}
Choosing an optimization algorithm for the DVRP
on the basis of the~characteristics of initial requests sets
leads to the~improvement of the~results.
Choosing between MEMSO and 2MPSO algorithms resulted
in~a~0.6\% improvement in comparison with the 2MPSO average performance
(with a maximum of 1.5\% improvement if all the predictions were accurate)
and 2.8\% in comparison with the MEMSO performance.
The best performance has been improved by 0.2\% on average
in comparison with 2MPSO and by 2.8\% in comparison with MEMSO.

The obtained results suggest that the proper benchmark features
has been selected and choosing an optimization algorithm
for a given benchmark problem is possible and may lead
to results improvement. Both the spatial and the volume related
features have been marked as significant in predicting relative
performance.

The future work should include an algorithm choice possibility
during the optimization and not only at the beginning of the working day.
It might be also beneficial to search for another set of characteristics
allowing for a proper choice of the algorithm, in order to try eliminating
the DVRP tasks for which the significantly worse algorithm has been selected.

\section*{Acknowledgments}
The research was financed by the~National Science Centre
in Poland grant number DEC-2012/07/B/ST6/01527\\
Project website: \url{http://www.mini.pw.edu.pl/~mandziuk/dynamic}.

\bibliographystyle{abbrv}
\bibliography{HyperDVRP,HyperDVRPxref}

\begin{thebibliography}{10}

\bibitem{burke2003hyper}
E.~Burke, G.~Kendall, J.~Newall, E.~Hart, P.~Ross, and S.~Schulenburg.
\newblock Hyper-heuristics: An emerging direction in modern search technology.
\newblock {\em International series in operations research and management
  science}, pages 457--474, 2003.

\bibitem{VRP:ChristofidesBeasley84}
N.~Christofides and J.~E. Beasley.
\newblock The period routing problem.
\newblock {\em Networks}, 14(2):237--256, 1984.

\bibitem{PSO:Convergence}
C.~W. Cleghorn and A.~P. Engelbrecht.
\newblock Particle swarm variants: standardized convergence analysis.
\newblock {\em Swarm Intelligence}, 9(2-3):177--203, 2015.

\bibitem{PSO:Params}
I.~Cristian and Trelea.
\newblock The particle swarm optimization algorithm: convergence analysis and
  parameter selection.
\newblock {\em Information Processing Letters}, 85(6):317--325, 2003.

\bibitem{TSP:2OPT}
G.~Croes.
\newblock A method for solving traveling salesman problems.
\newblock {\em Operations Res. 6}, pages 791--812, 1958.

\bibitem{VRP:Definition}
G.~B. Dantzig and R.~Ramser.
\newblock {The Truck Dispatching Problem}.
\newblock {\em Management Science}, 6:80--91, 1959.

\bibitem{DVRP:ACOLNS}
M.~Elhassania, B.~Jaouad, and E.~A. Ahmed.
\newblock A new hybrid algorithm to solve the vehicle routing problem in the
  dynamic environment.
\newblock {\em International Journal of Soft Computing}, 8(5):327--334, 2013.

\bibitem{DVRP:GA2014}
M.~Elhassania, B.~Jaouad, and E.~A. Ahmed.
\newblock Solving the dynamic vehicle routing problem using genetic algorithms.
\newblock In {\em Logistics and Operations Management (GOL), 2014 International
  Conference on}, pages 62--69. IEEE, 2014.

\bibitem{VRP:FisherJakumar81}
M.~L. Fisher and R.~Jaikumar.
\newblock A generalized assignment heuristic for vehicle routing.
\newblock {\em Networks}, 11(2):109--124, 1981.

\bibitem{DVRP:GA:TS}
F.~T. Hanshar and B.~M. Ombuki-Berman.
\newblock Dynamic vehicle routing using genetic algorithms.
\newblock {\em Applied Intelligence}, 27(1):89--99, Aug. 2007.

\bibitem{PSO:Introduction}
J.~Kennedy and R.~Eberhart.
\newblock {Particle Swarm Optimization}.
\newblock {\em {Proceedings of IEEE International Conference on Neural
  Networks. IV}}, pages 1942--1948, 1995.

\bibitem{DVRP:MAPSO}
M.~R. Khouadjia, E.~Alba, L.~Jourdan, and E.-G. Talbi.
\newblock {Multi-Swarm Optimization for Dynamic Combinatorial Problems: A Case
  Study on Dynamic Vehicle Routing Problem}.
\newblock In {\em Swarm Intelligence}, volume 6234 of {\em Lecture Notes in
  Computer Science}, pages 227--238. Springer, Berlin / Heidelberg, 2010.

\bibitem{DVRP:DAPSO}
M.~R. Khouadjia, B.~Sarasola, E.~Alba, L.~Jourdan, and E.-G. Talbi.
\newblock {A comparative study between dynamic adapted PSO and VNS for the
  vehicle routing problem with dynamic requests}.
\newblock {\em Applied Soft Computing}, 12(4):1426--1439, 2012.

\bibitem{DVRP:MEMSO}
M.~R. Khouadjia, E.-G. Talbi, L.~Jourdan, B.~Sarasola, and E.~Alba.
\newblock Multi-environmental cooperative parallel metaheuristics for solving
  dynamic optimization problems.
\newblock {\em Journal of Supercomputing}, 63(3):836--853, 2013.

\bibitem{DVRP:Study}
P.~Kilby, P.~Prosser, and P.~Shaw.
\newblock {Dynamic VRPs: A Study of Scenarios}, 1998.

\bibitem{DVRP:Ants}
R.~Montemanni, L.~Gambardella, A.~Rizzoli, and A.~Donati.
\newblock A new algorithm for a dynamic vehicle routing problem based on ant
  colony system.
\newblock {\em Journal of Combinatorial Optimization}, 10:327--343, 2005.

\bibitem{DVRP:2PSO}
M.~Okulewicz and J.~Ma{\'n}dziuk.
\newblock {Application of Particle Swarm Optimization Algorithm to Dynamic
  Vehicle Routing Problem}.
\newblock In L.~Rutkowski, M.~Korytkowski, R.~Scherer, R.~Tadeusiewicz,
  L.~Zadeh, and J.~Zurada, editors, {\em Artificial Intelligence and Soft
  Computing}, volume 7895 of {\em Lecture Notes in Computer Science}, pages
  547--558. Springer Berlin Heidelberg, 2013.

\bibitem{DVRP:2MPSO}
M.~Okulewicz and J.~Ma{\'n}dziuk.
\newblock {Two-Phase Multi-Swarm PSO and the Dynamic Vehicle Routing Problem}.
\newblock In {\em 2nd IEEE Symposium on Computational Intelligence for
  Human-like Intelligence}, pages 86--93, Orlando, Fl, USA, 2014. IEEE Press.

\bibitem{DVRP:MCTree}
M.~Okulewicz and J.~Ma{\'n}dziuk.
\newblock {Dynamic Vehicle Routing Problem: A Monte Carlo approach}.
\newblock In {\em Information Technologies: Research and Their
  Interdisciplinary Applications 2015}, pages 119--138, Pu{\l}tusk, Poland,
  2015.
  \url{http://phd.ipipan.waw.pl/pliki/mat_konferencyjne/12_ITRIA_2015_01.pdf#page=120}.

\bibitem{DVRP:Review2011}
V.~Pillac, M.~Gendreau, C.~Gu\'{e}ret, and A.~L. Medaglia.
\newblock A review of dynamic vehicle routing problems.
\newblock {\em European Journal of Operational Research}, 225(1):1--11, 2013.

\bibitem{Psaraftis1980130}
H.~N. Psaraftis.
\newblock Dynamic programming solution to the single vehicle many-to-many
  immediate request dial-a-ride problem.
\newblock {\em Transportation Science}, 14(2):130--154, 1980.
\newblock cited By 198.

\bibitem{DVRP:Reintroduction}
H.~N. Psaraftis.
\newblock Dynamic vehicle routing: Status and prospects.
\newblock {\em Annals of Operations Research}, 61(1):143--164, 1995.

\bibitem{PSO:Modified}
Y.~Shi and R.~Eberhart.
\newblock A modified particle swarm optimizer.
\newblock {\em Proceedings of IEEE International Conference on Evolutionary
  Computation}, pages 69--73, 1998.

\bibitem{VRP:Taillard93}
{\'E}.~D. Taillard.
\newblock Parallel iterative search methods for vehicle routing problems.
\newblock {\em Networks}, 23(8):661--673, 1993.

\end{thebibliography}

\end{document}